\documentclass[conf]{new-aiaa}
\usepackage[utf8]{inputenc}
\usepackage[version=4]{mhchem}
\usepackage{ifpdf}
\usepackage[pdftex]{graphicx}
\usepackage{algorithm}
\usepackage[noend]{algpseudocode}
\usepackage{array}
\usepackage{subcaption}
\usepackage{url}
\usepackage{multicol}
\usepackage{booktabs}
\usepackage{siunitx}
\usepackage{color}

\usepackage{amsmath,amsfonts}
\usepackage{rotating}
\usepackage{enumerate}

\usepackage{caption}
\usepackage{wrapfig}
\usepackage[graphicx]{realboxes}

\usepackage{stfloats}
\usepackage{float}

\usepackage{adjustbox}
\usepackage{textcomp}
\usepackage{xcolor}

\title{Path Planning for Optimal Coverage of Areas with Nonuniform Importance}

\author{Gregory Snyder\footnote{Graduate Student, Department of Mechanical Engineering, snyderg@hawaii.edu.}, Sachin Shriwastav\footnote{Graduate Student, Department of Mechanical Engineering, sachins@hawaii.edu. (Corresponding author)}, Dylan Morrison-Fogel\footnote{PhD Graduate, Department of Mechanical Engineering.} and Zhuoyuan Song\footnote{Assistant Professor, Department of Mechanical Engineering, zsong@hawaii.edu, AIAA Member.\\\indent \indent (Snyder and Shriwastav contributed equally to this work.)}}
\affil{University of Hawai`i at M\=anoa, Honolulu, HI 96822}

\begin{document}

\maketitle

\begin{abstract}
\label{Abstract}

Coverage of an inaccessible or challenging region with potential health and safety hazards, such as in a volcanic region, is difficult yet crucial from scientific and meteorological perspectives. Areas contained within the region often provide valuable information of varying importance. We present an algorithm to optimally cover a volcanic region in Hawai`i with an unmanned aerial vehicle (UAV). The target region is assigned with a nonuniform coverage importance score distribution. For a specified battery capacity of the UAV, the optimization problem seeks the path that maximizes the total coverage area and the accumulated importance score while penalizing the revisiting of the same area. Trajectories are generated offline for the UAV based on the available power and coverage information map. The optimal trajectory minimizes the unspent battery power while enforcing that the UAV returns to its starting location. This multi-objective optimization problem is solved by using sequential quadratic programming. The details of the competitive optimization problem are discussed along with the analysis and simulation results to demonstrate the applicability of the proposed algorithm.

\end{abstract}

\section{Introduction}
\label{Introduction}

Over the last century, the Hawai`i Volcanic Observatory (HVO) has developed volcano-monitoring systems and networks to record and document activities at Hawaiian volcanoes. The existing infrastructure includes over a hundred field stations with instruments that record and measure earthquakes, ground movement, volcanic gases, sound waves, lava advancement, magma volume below ground, and visual changes in eruptive activities. Satellite data have been used to detect changes in ground elevation and surface temperature, which can indicate lava or other eruptive activities, however direct line-of-sight is often interrupted with cloud cover and volcanic haze. Without satellite coverage, the large surface area of Hawai`i island presents issues for remote sensing, therefore implementation of autonomous unmanned aerial vehicles (UAVs) may prove invaluable for HVO. Given the size of commercially available UAVs, power restrictions limit the total surface area that a single platform may observe. Therefore, an optimization scheme must be implemented to determine the most effective battery configuration of UAV for a specific environment and the optimal path for that specific UAV to take based on this configuration. 

Path planning algorithms for UAV systems have been considered for a variety of applications, though a majority examine a problem of reaching the greatest number of targets while expending minimal energy or time~\cite{SonmezeKocyigitKugu, GeigerHornDeLulloNiessnerLong, SujitBeard}. Voronoi diagrams have been traditionally implemented for graphical methods \cite{Bortoff,Dogan,ButenkoMurpheyParadalos}, while spline generation and single-objective genetic algorithms are becoming commonplace for both online and offline operations \cite{NikolosValavantisTsourveloudisKostaras, ArantesArantesToledoWilliams}. The means by which the cost of a system is represented varies considerably; for example, Bortoff \cite{Bortoff} considers both a two-step optimization algorithm that first explores graphical search methods and secondly generates an equilibrium path solution for a Lagrangian mechanical system composed of virtual forces. Combinations of graphical and mechanical approaches provide a means of reducing a solution search area prior to processing higher-order system dynamics.

In general, path planning has been used to achieve UAV interaction with specific targets or to maximize coverage over a specific area of uniform importance. A natural succession is to maximize coverage over a given area, within which some known locations are assigned higher importance than others. Topological approaches have been investigated by Li et al. \cite{LiWangSun} that address energy maps for UAV paths along variable topological terrain. This approach considers energy-optimal area coverage under the specific dynamics of the UAV, but does not consider the possibility of a performance index being nonuniform along identical topological levels, that is, regions of greater importance to data collection. The problem that arises for nonuniform area preference, discussed by Mittal and Deb \cite{MittalDeb}, is the normalization of the objective function under the influence of a weight vector. This consideration implies that a topological map should be generated for each performance index considered, as well as the weight vector at every iteration.

This work intends to explore the coverage optimization problem where power limitation and nonuniform coverage importance distribution are considered. An optimization problem will be formulated and solved for a UAV deployed over the area, with an objective to maximize the area covered, energy utilization and accumulated coverage importance. The major contributions of the proposed algorithm are: (i) Optimized coverage of an area for given power for a UAV, by maximizing the coverage and minimizing the return energy, (ii) Trade-off between coverage of new area and net information gain, by attempting to cover areas of higher coverage importance while penalizing revisits, (iii) simple yet effective coverage algorithm readily applicable to real-world applications.

The remainder of the paper has been organized as follows. Section~\ref{ProblemDefinition} defines and sets up the problem, and Section~\ref{ProposedApproach} discusses the optimization problem formulation and the proposed approach. Preliminary results and analysis are discussed in Section~\ref{PreliminaryResults}, and Section~\ref{conclusion} concludes the paper and lists some future research directions.

\begin{figure}[]
    \centering
    \includegraphics[width=0.5\linewidth]{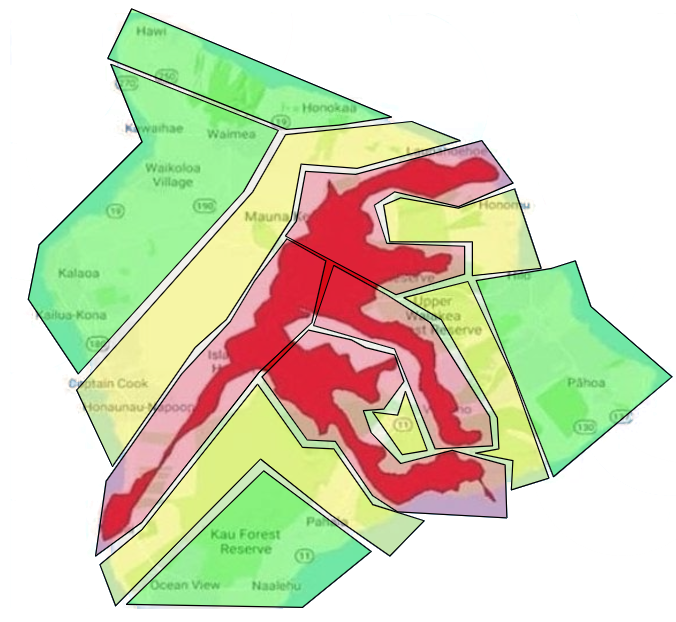}
    \caption{An example volcanic survey area (Hawai`i Island), with the overlaid coverage importance distribution (Red: High, Yellow: Medium and Green: Low)}
    \label{Fig:Illustration}
\end{figure}

\section{Problem Definition}
\label{ProblemDefinition}

Considering the problem of monitoring volcanic activities with a UAV, we wish to maximize the observation productivity by varying its battery capacity as well as its flight path design. Observation productivity is defined as the norm of a performance index encoding the area coverage, the number of ``waypoints", and the unspent energy upon return. The problem does not consider real-time path planning, rather a prescribed flight plan. To best solve our problem we need the UAV to maximize its surveyed area, while covering more important areas and best utilize its limited power before returning to the base station. 

\begin{figure*}[]
    \centering
    \includegraphics[width=0.75\textwidth]{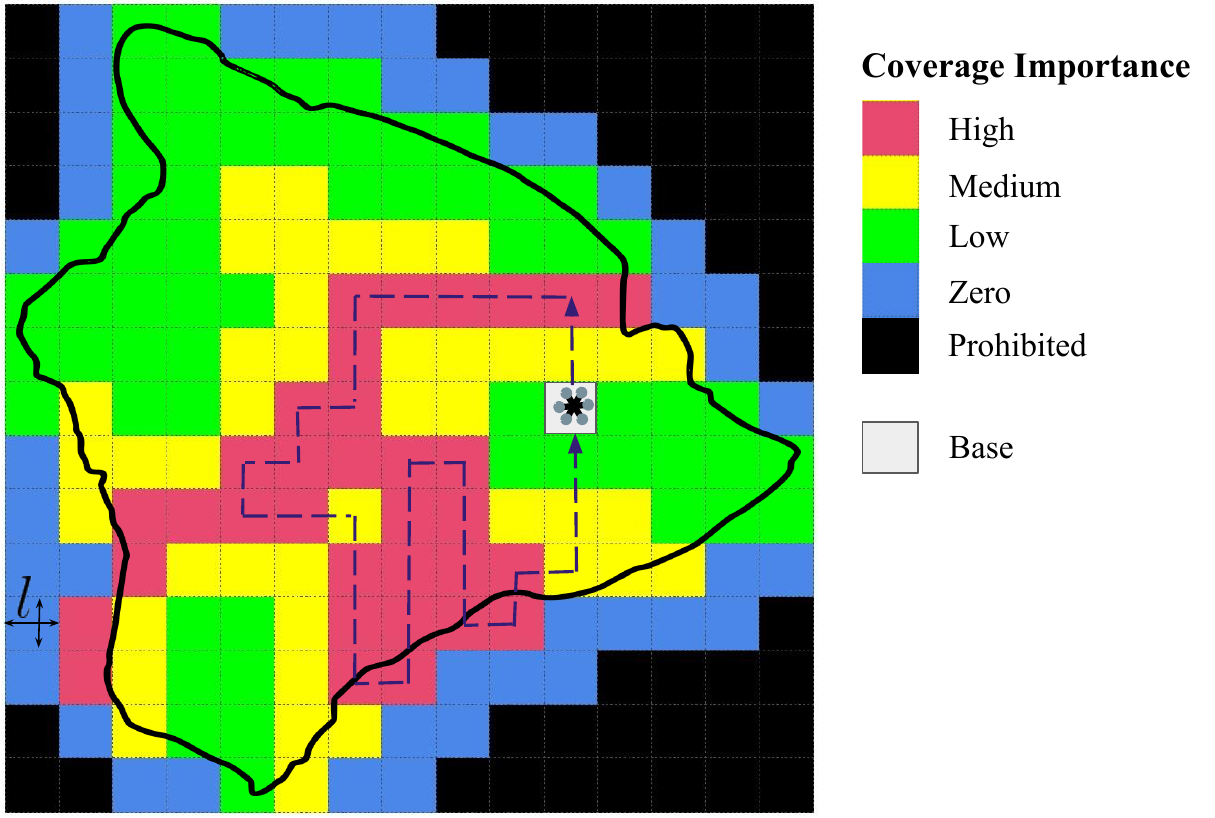} 
    \caption{Coverage importance distribution grid overlaid on the desired coverage area. An example trajectory to maximize coverage and net area information gain with a UAV starting from and returning to the base is also shown.}
    \label{Fig:Gridpath}
\end{figure*}

We assume that the coverage importance map (See Fig.~\ref{Fig:Illustration}) is available for trajectory planning. The problem is then set up by defining a grid over the target area using the area boundary information. The grid is then created at the spacing $l$ in both directions of the area, and the undesired coverage locations outside the area are then identified and assigned with the lowest importance weights. An example where the coverage importance map is overlaid on top of the area map is shown in Fig.~\ref{Fig:Gridpath}. The optimal path planning is carried out as discussed in the following section.

\section {Proposed Approach}
\label{ProposedApproach}

\begin{figure*}[]
    \centering
    \includegraphics[width=0.55\textwidth]{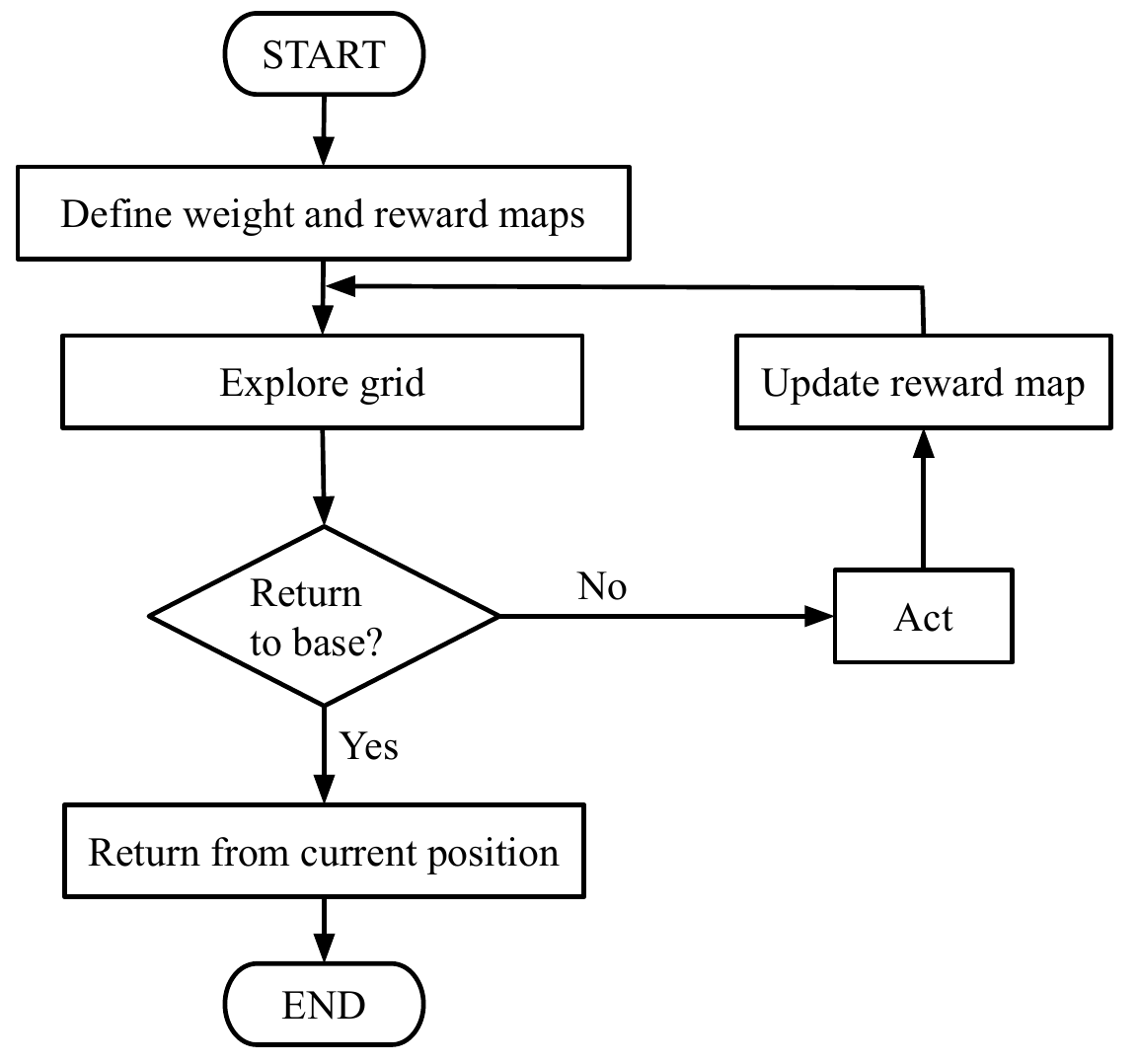} 
    \caption{Flowchart for the proposed optimal path planning algorithm.}
    \label{Fig:Flowchart}
\end{figure*}

The proposed approach to maximize the coverage over a volcanic area and net coverage information gain is outlined in Algorithm~\ref{alg1}, and the path planning logic is shown in Fig.~\ref{Fig:Flowchart}. The boundary information of the area is used to create a grid over it, and the coverage information importance map is overlaid. The UAV starts at a prescribed base station ($x_\text{base}, \ y_\text{base}$) as shown by a white square in Fig.~\ref{Fig:Gridpath}, and it has the knowledge of the overall grid, its own battery level, and the cost to move in each possible direction on the grid. For this work, we assume that the cost to move is same in each direction. The algorithm also maintains the history of the visited locations to discount the trajectories containing repetitions. However, we do not prohibit revisits since they may be necessary as part of the shortest return path to the base due to battery level or physical map constraints. The coverage of new areas thus competes with the net coverage information gain and the battery utilization in the optimization problem. At each time step, the UAV lists the candidate locations to move, all of which are $l$ units away. During this process, the UAV keeps a record of the cells covered and keeps track of its battery level relative to its current distance from the base. The UAV is programmed to initiate the return to the base when the battery level is critical for safe return from its present position while still attempting to maximizing its total coverage on its safe return.

When the UAV starts up it will first determine a target area to reach by weighing the reward for reaching an area based on its importance, as well as by a temporary reward map. The weight map corresponds to the importance of a specific cell. 
The reward map will be repeatedly generated as seen in Fig.~\ref{Fig:Flowchart}, where the value of each cell is determined by the importance weight of the cell and penalized by its distance relative to the UAV's current position as well whether the cell has been previously visited.
Once a reward map is generated, the UAV will calculate the shortest path to the target cell of the highest reward using Djikstra's algorithm ~\cite{dijkstra1959note}, where the edge length is the reward of the cell. While this does not necessarily yield the shortest path to each following target point, it will generate the shortest path that yields a positive reward and and maximize the accumulated reward along this path, provided it has unspent energy to do so. Once the path to the target cell is calculated, it will then calculate the shortest return path to the base using Djikstra's algorithm ~\cite{dijkstra1959note} and Bellman-Ford algorithm~\cite{bellman1958routing, moore1959shortest, ford1956network} from the candidate target position by choosing the shortest path to base, without possibly revisiting cells to maximize the total reward. The UAV proceeds to the next exploration step only if the remaining battery level is higher than the threshold requirement for the calculated shortest path to base. This is repeated at every new location to fully use the available power.

\begin{figure*}[t]
	\centering
	\begin{subfigure}[b]{0.32\textwidth}
	\includegraphics[width=1\linewidth]{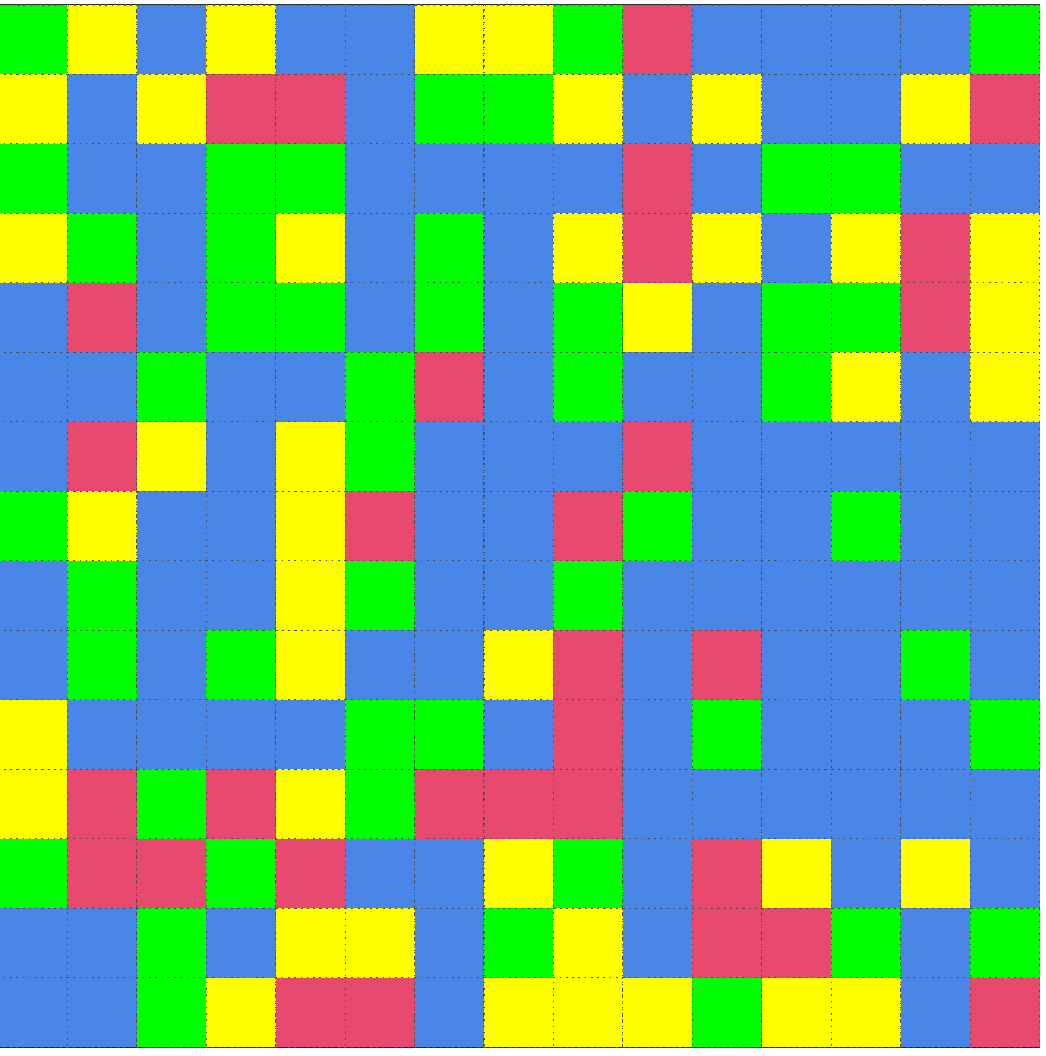}
	\caption*{(a)}
	\end{subfigure}
	\hspace{1mm}
	\begin{subfigure}[b]{0.32\textwidth}
    \includegraphics[width=1\linewidth]{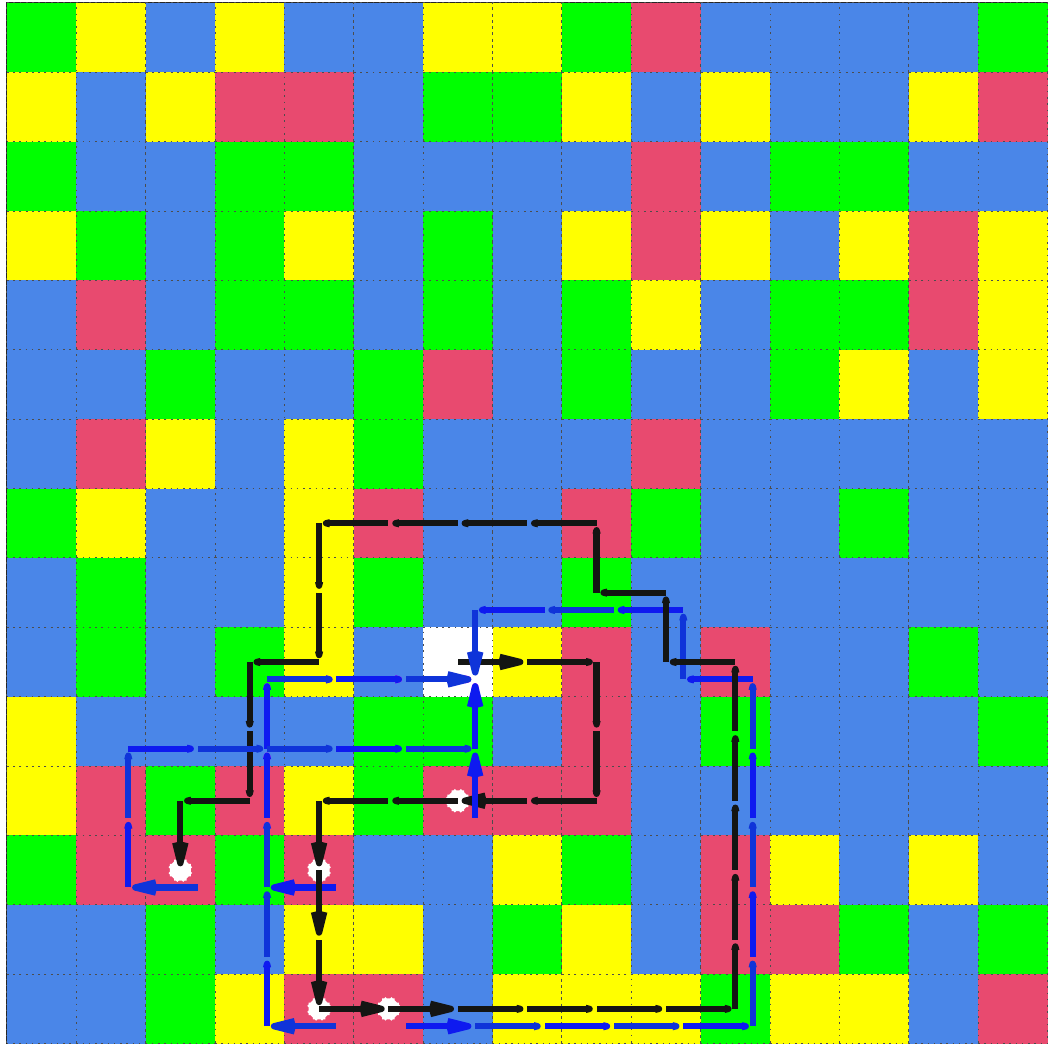}	
    \caption*{(b)}
	\end{subfigure}
	\hspace{1mm}
	\begin{subfigure}[b]{0.32\textwidth}
    \includegraphics[width=1\linewidth]{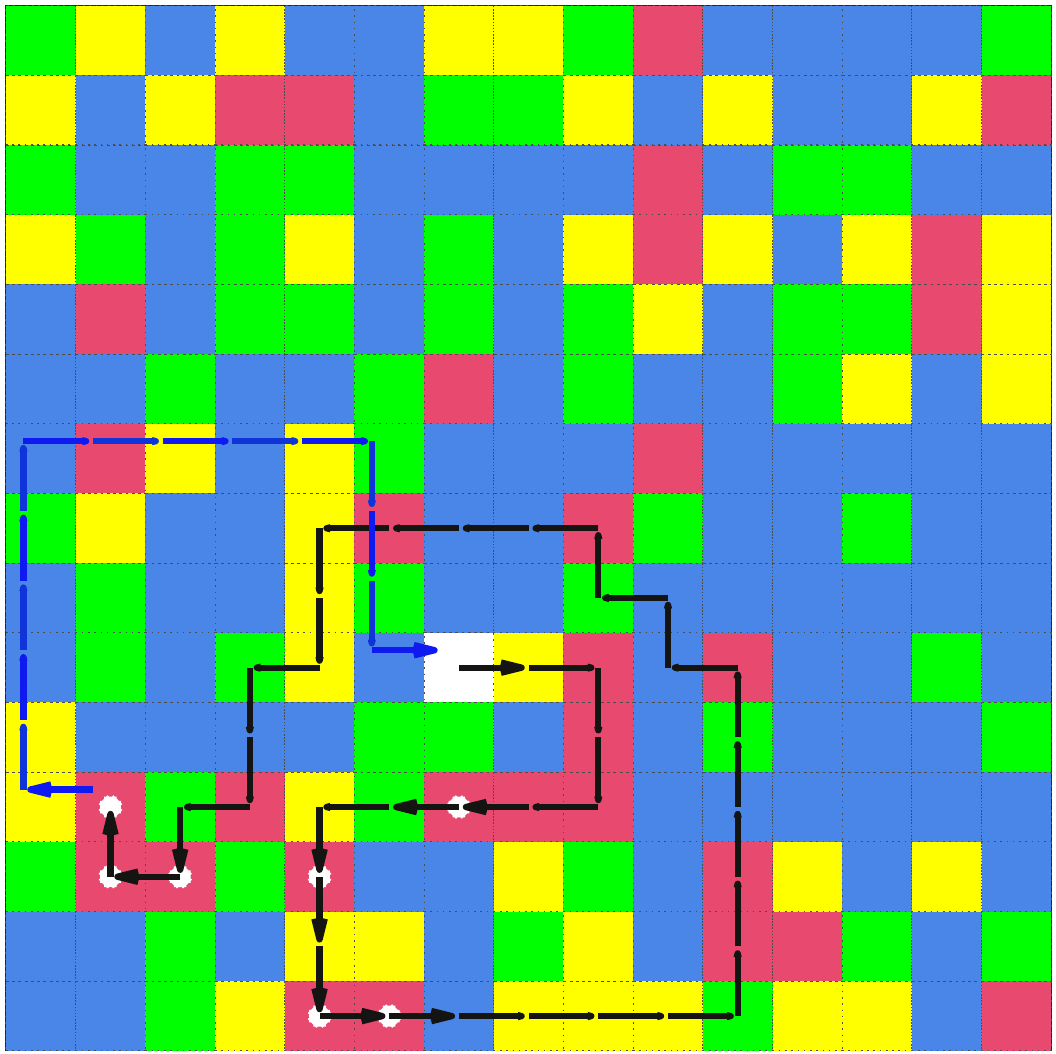}	
    \caption*{(c)}
	\end{subfigure}
    \caption{Illustration of the development of the path planning algorithm. The red, yellow, green and blue cells denote the highest, medium, low and lowest reward for covering the respective areas respectively. The white cell denotes the base location and the white dots show the next location of highest importance that the algorithm targets based on the reward values of the cells and its current distance from the UAV. The outgoing and the return trajectories have been shown in black and blue, respectively. The figure shows (a) randomized weight map, (b) five next highest importance locations, their outgoing and respective instantaneous return trajectories, and (c) final trajectory.}
	\label{fig:presults}
\end{figure*}

Fig.~\ref{fig:presults} illustrates how the algorithm would navigate a map of pseudo-randomly distributed cells with weights of high, medium, low and zero values marked in red, yellow, green, and blue, respectively.
The white cell denotes the base location where the UAV starts its path planning and ultimately must return to.
The white dots show the target waypoint with the highest expected reward calculated from a dynamic weight map that the algorithm maintains based on the importance of the cells, their distances from the UAV, and if it has previously visited.
As seen in Fig.~\ref{fig:presults}(b), the next trajectory is more affected by the distance to the target rather than the energy expense due to the large difference in the anticipated reward of the cells the path covers.
Despite this the optimal path will take into account the reward of visiting more importance cells against the cost of moving to an additional cell. 
Trajectories in these circumstances can be altered by adjusting the penalties of revisiting a cell. 
The black arrows denote the outgoing trajectory from each cell and the blue arrows denote what the algorithm deems to be the the shortest-path of return with the maximum reward based on the current remaining power.
These shortest paths with maximum reward are calculated at every target, as shown in the figure. 
These return trajectories are determined using Djikstra's algorithm depending on whether a possible trajectory passes over a visited space as the assigned penalties provide a higher penalty.
Finally, Fig.~\ref{fig:presults}(c) shows the complete exploration and return trajectory for the UAV for the example setup as the algorithm has determined a path that consumes all the remaining power from that waypoint.  

\begin{algorithm}[t!]
	\caption{Optimal Coverage of Area with Nonuniform Importance}
    \begin{algorithmic}[1]
		\Statex \textbf{Input:} Area grid, cell indices ($M$), coverage importance weight map ($\gamma(M)$), base coordinates ($x_\text{base}, \ y_\text{base}$), cell size ($l$), starting battery level ($B_\text{start}$), critical battery level ($B_\text{min}$)
		\Statex \textbf{Output:} Optimal trajectory $M_i \ (i=1,2, \ldots, n)$
		\Statex \textbf{Notations:} 
		\Statex \qquad $\bullet$ $D$ = Grid matrix listing for visited ($\nu = 1$) and not visited ($\nu = 0$)
		\Statex \qquad $\bullet$ $B_i$ = Current battery level
		\Statex \qquad $\bullet$ $B_\text{min}$ = Critical battery level required to return to base from current position
		\Statex \qquad $\bullet$ $B_{\text{target},j}$ = Energy cost to move to target, $j$
		\Statex \qquad $\bullet$ $W_i$ = Generated reward map from current location
        \medskip 
		\State Initialize reward map $W_1$
		\State Start at ($x_\text{base}, \ y_\text{base}$)
		\State $i=1$
		\While {$B_i$ > 0}
		    \State Update $W_i$ for available options where $W_i = \gamma(M_i) -\|M_{k}-M_i\| \ \forall \ k \ne i$
		    \State $M_{i+1} = M^*_{i+1} \triangleq \underset{M_{i+1}}{\mathrm{argmin}} \ W_i(\gamma(M_{i+1}))$ \Comment{(\ref{eqn:J1}), (\ref{eqn:J3})} \label{r2t}
		    \State Update $W_i(\cdot)$ with $D_i$
		    \State Calculate to path to ($x_\text{base}, \ y_\text{base}$) from $\gamma(M_{i})$ \label{r2b}
		    \State Calculate $B_\text{min}$ 
		    \If{$B_i > B_{\text{target},j} + B_{\text{min},j}$}
		        \State Move to $M^{*}_{i}$ via path generated in Line ~\ref{r2t} 
	        \EndIf
	       \State $i$++
            
		\EndWhile
		  	\State Return to base using path generated in Line~\ref{r2b} from $M_{i-1}$
    \end{algorithmic}
	\label{alg1}
\end{algorithm}

\section{Formulating the optimization problem}

In this work we formulate an optimization problem comprised of three objectives. The performance indices for the proposed approach are total coverage ($J_1$), unspent energy ($J_2$), and the net coverage importance gain ($J_3$).
To maximize the coverage, $J_1$, we have the cost term:
\begin{equation} \label{eqn:J1}
     J_1= l^2 \sum_{i=1}^n \nu_i(M_i) ,    
\end{equation}
where $i \in \{ 1, \ldots, n \}$ is the waypoint index, $l$ is size of the cell, $M_i \triangleq (x_i,y_i)$ is the cell occupied by waypoint $i$, and $\nu_i$ is a Boolean function that equals 0 when waypoint $i$ has been accounted for and $1$ otherwise.. 

To minimize remaining energy upon return, the second cost term may be stated as
\begin{equation}  \label{eqn:J2}
    J_2 = B_{\text{start}} -\sum_{j=1}^m B_{\text{target},j} - B_{\text{final}}
\end{equation}
where $B_\text{start}$ is the initial battery level at deployment, $j \in \{ 1, \ldots, m \}$ where $j$ is the current target, $B_\text{target}$ is the energy cost to move to the target from the current cell and $B_\text{final}$ is the cost of returning to base from the final target waypoint. $J_2$ keeps a check on the total energy left during the trajectory planning, so as to use the available power to the maximum, while trying to cover as many new cells as possible. It also keeps track of the power left at a given waypoint to make sure the UAV has enough power left during its return to visit the maximum possible uncovered waypoints, instead of having to trace a shorter return path and overlap with the already covered points. 

To maximize the allotted way point values to favor the covered area to specific targets, we will add the following cost term:
\begin{equation}  \label{eqn:J3}
    J_3= l^2 \sum_{i=1}^n \gamma_i(M_i) \nu_i(M_i) .
\end{equation}
Along with other variables explained in $J_1$, $\gamma_i$ is the function relating the reward for the cell at $M_i$; the higher value of which means the point is expected to cover some crucial information and it is beneficial to visit it.
This term is unique from $J_1$ in that if needed we can increase the weights of visiting new locations to counter higher weighted reward cells. 
$J_3$ allows us to minimize the coverage importance gain, and dictates the likely order of the area coverage.
The resulting cost function for the resulting minimization problem becomes,
\begin{equation}  \label{eqn:J_tot}
    J_\text{total} = - \alpha_1 J_1 + \alpha_2 J_2 - \alpha_3 J_3,
\end{equation}
where  $\alpha_1, \ \alpha_2 \text { and } \alpha_3$ are the tunable weights imposed on $J_1$, $J_2$ and $J_3$ respectively.

\section{Simulation Results}
\label{PreliminaryResults}

\begin{figure*}[]
	\centering
	\begin{subfigure}[b]{0.48\textwidth}
	\includegraphics[width=1\linewidth]{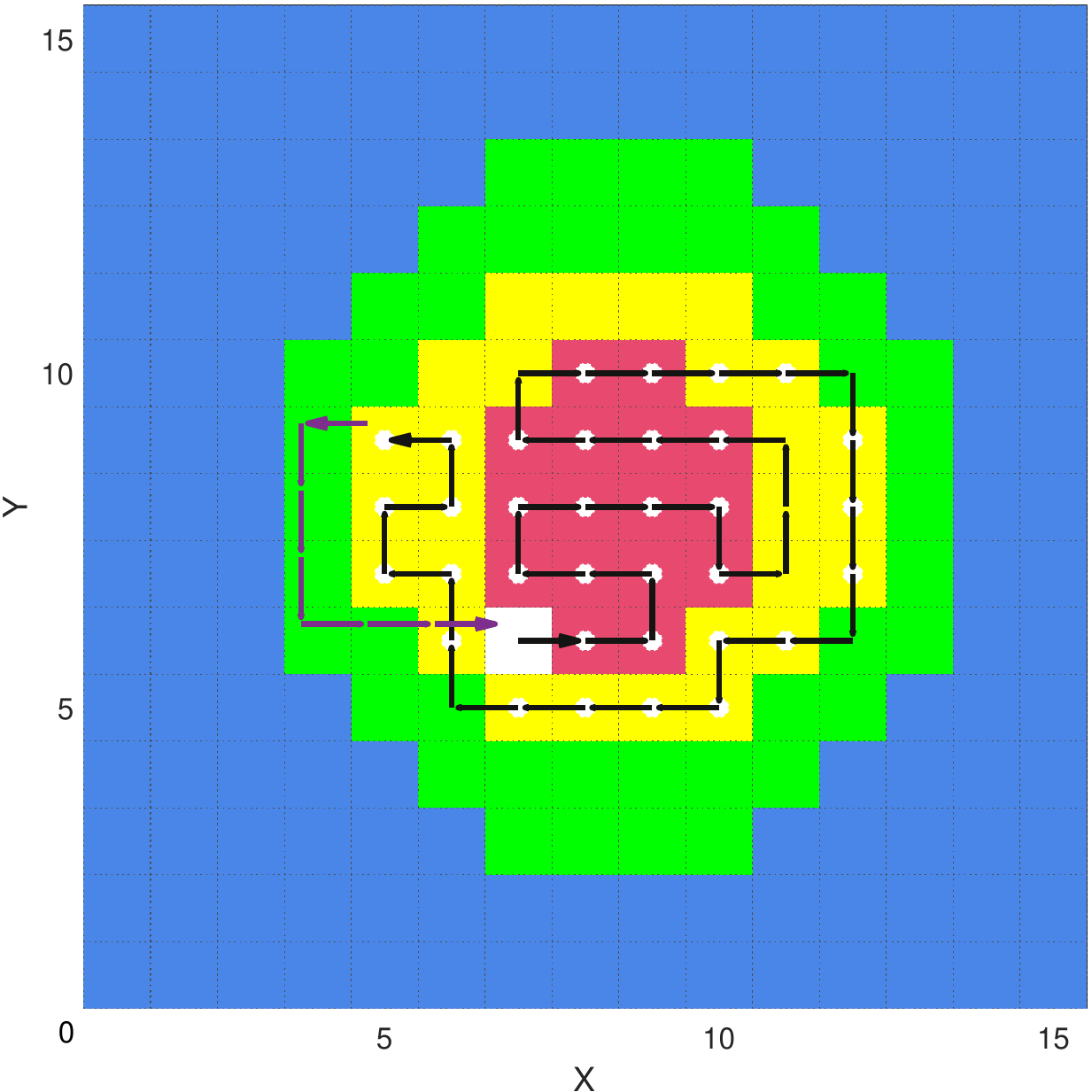}
	\caption*{\textnormal{(a) $[x_\text{base}, y_\text{base}] = [7, 6]$, $B_\text{start} = 50, \ J_2 = 4$}}
	\end{subfigure}
	\hspace{5mm}
	\vspace{5mm}
	\begin{subfigure}[b]{0.48\textwidth}
    \includegraphics[width=1\linewidth]{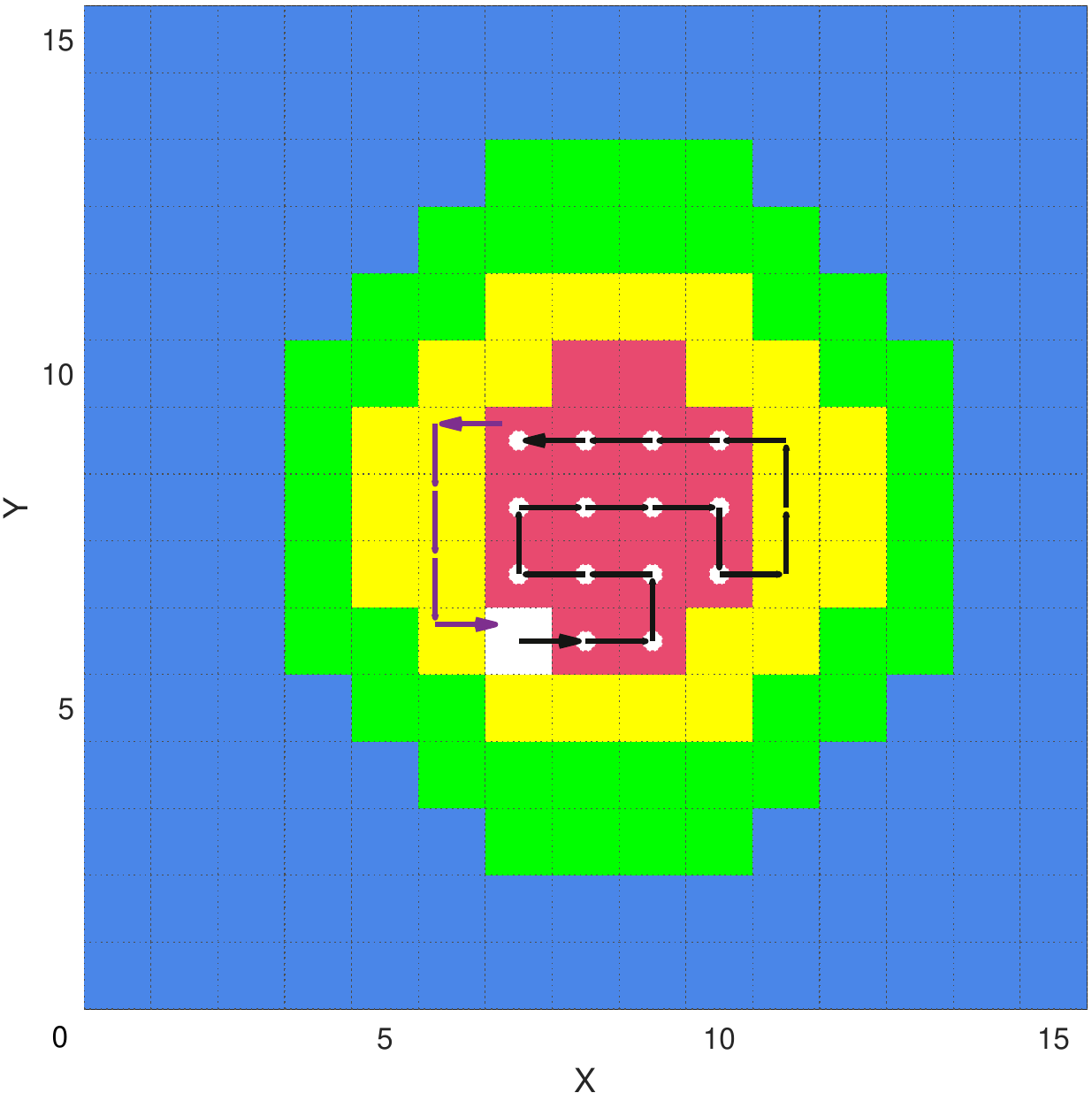}	
    \caption*{\textnormal{(b) $[x_\text{base}, y_\text{base}] = [7, 6]$, $B_\text{start} = 25, \ J_2 = 3$}}
	\end{subfigure}
	\vspace{5mm}
	\begin{subfigure}[b]{0.48\textwidth}
    \includegraphics[width=1\linewidth]{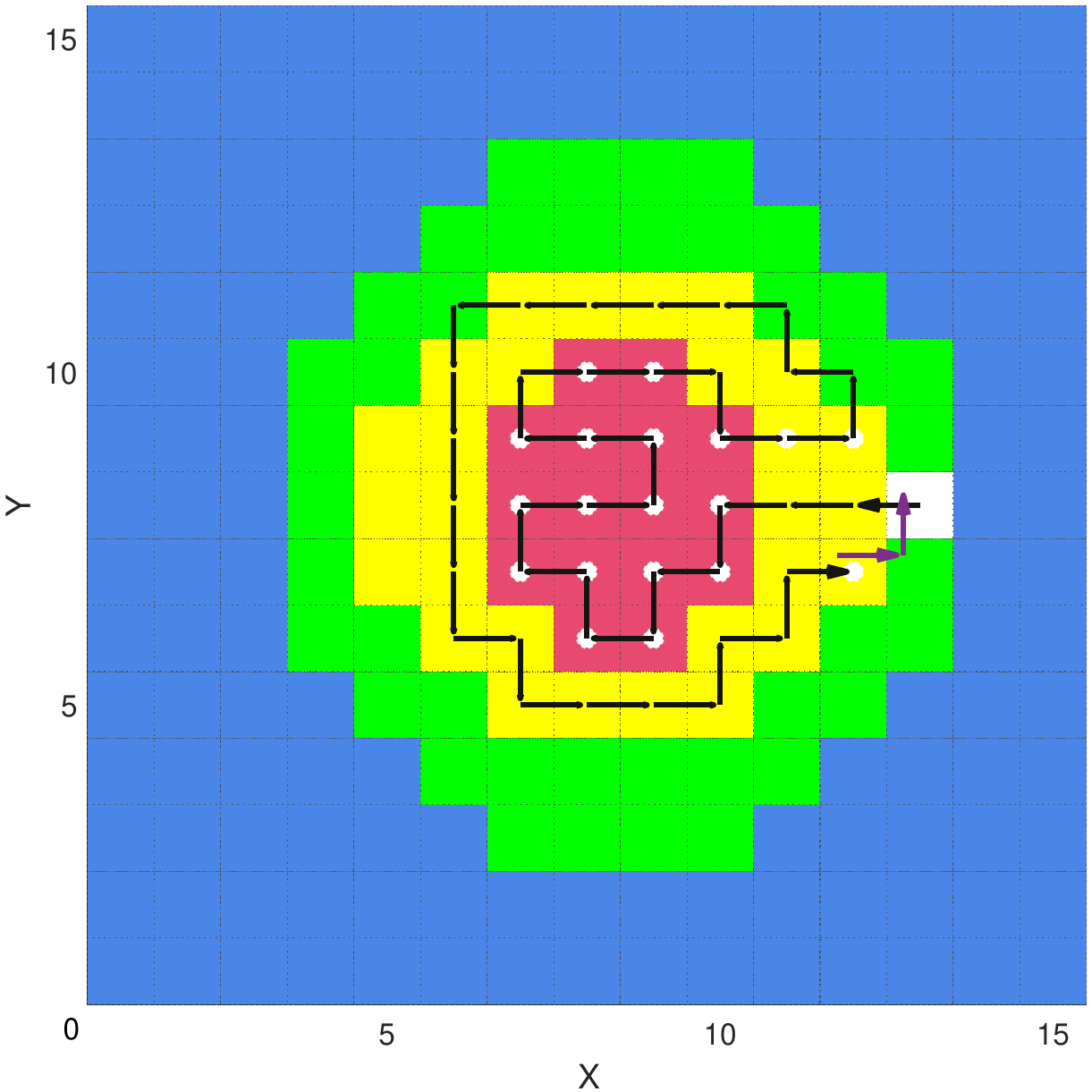}	
    \caption*{\textnormal{(c) $[x_\text{base}, y_\text{base}] = [13, 8]$, $B_\text{start} = 50, \ J_2 = 2$}}
	\end{subfigure}
	\hspace{5mm}
	\begin{subfigure}[b]{0.48\textwidth}
    \includegraphics[width=1\linewidth]{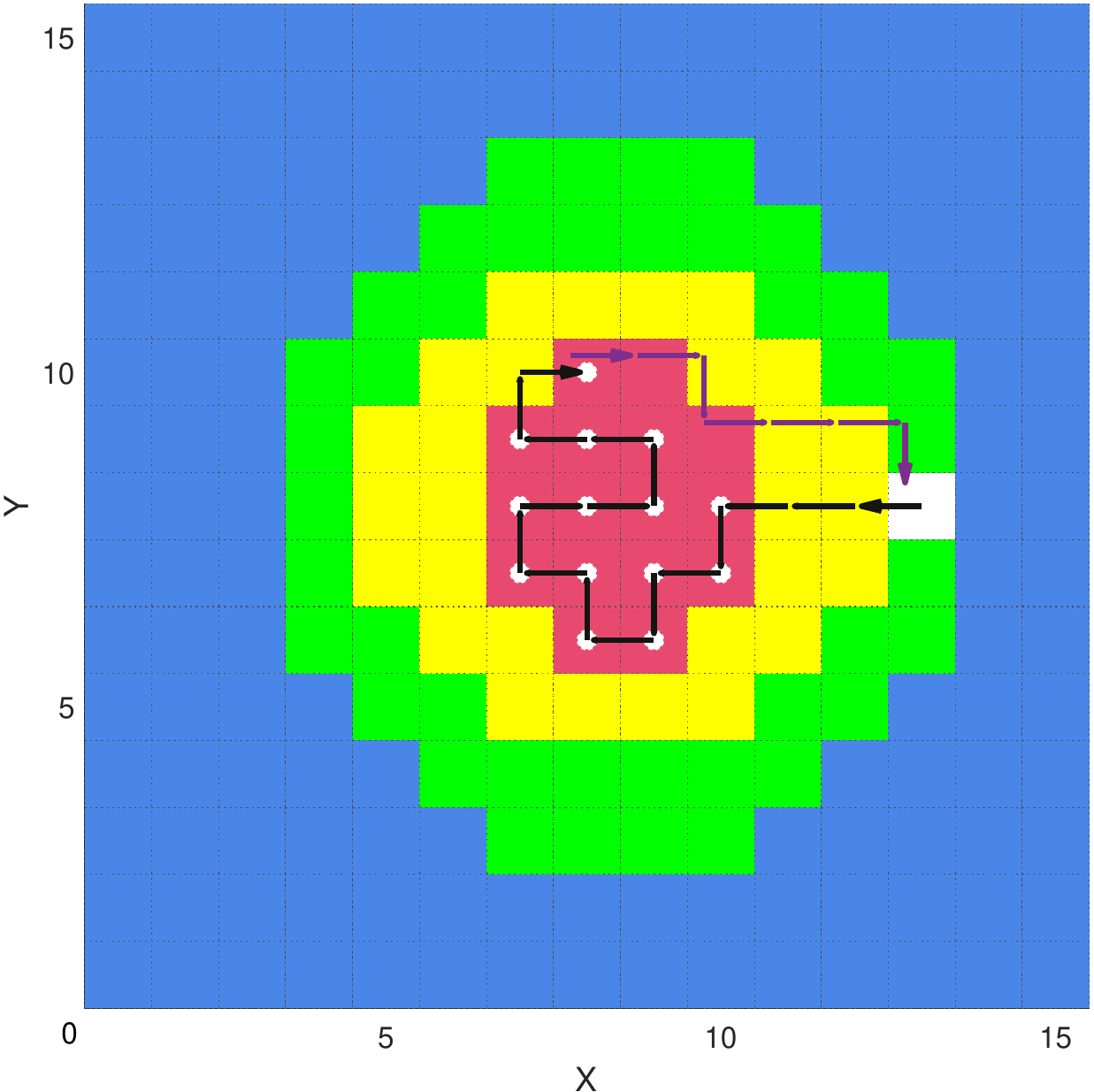}	
    \caption*{\textnormal{(d) $[x_\text{base}, y_\text{base}] = [13, 8]$, $B_\text{start} = 25, \ J_2 = 1$}}
	\end{subfigure}
    \caption{Application of the proposed algorithm on an example multivariate Gaussian distribution map for different initial locations and battery capacities. The red, yellow, green and blue cells denote the highest, medium, low and lowest coverage importance, respectively. The white cell denotes the base location and the white dots show the next instantaneous location with the highest importance. The black line denotes the outgoing trajectory, the purple line denotes the shortest-path return trajectory, and the arrowheads denote the directions.}
	\label{fig:GaussianTraj}
\end{figure*}

\begin{figure*}[]
	\centering
	\begin{subfigure}[b]{0.48\textwidth}
	\includegraphics[width=1\linewidth]{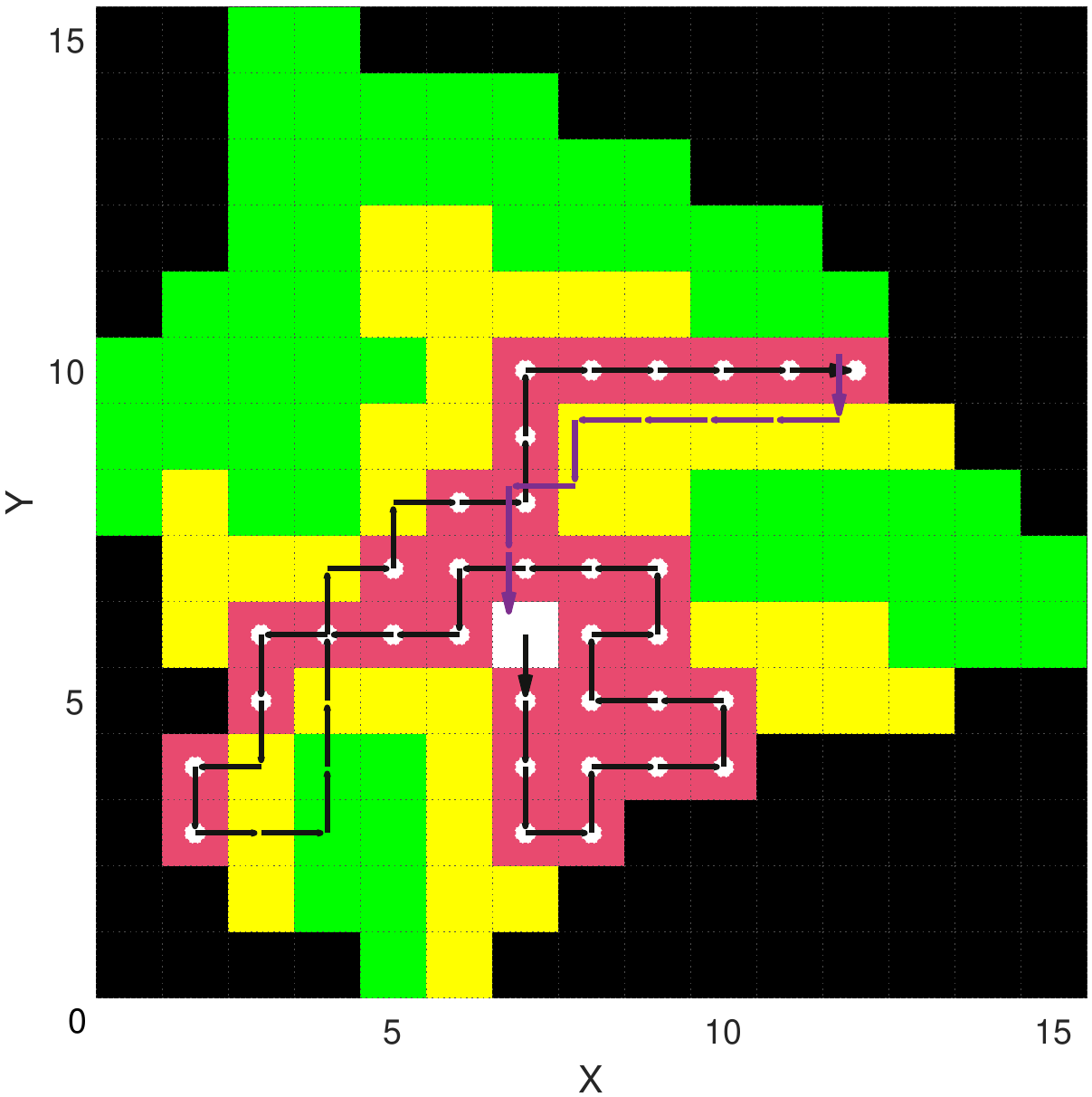}
	\caption*{\textnormal{(a) $[x_\text{base}, y_\text{base}] = [7, 6]$, $B_\text{start} = 50, \ J_2 = 0$}}
	\end{subfigure}
	\hspace{5mm}
	\vspace{5mm}
	\begin{subfigure}[b]{0.48\textwidth}
    \includegraphics[width=1\linewidth]{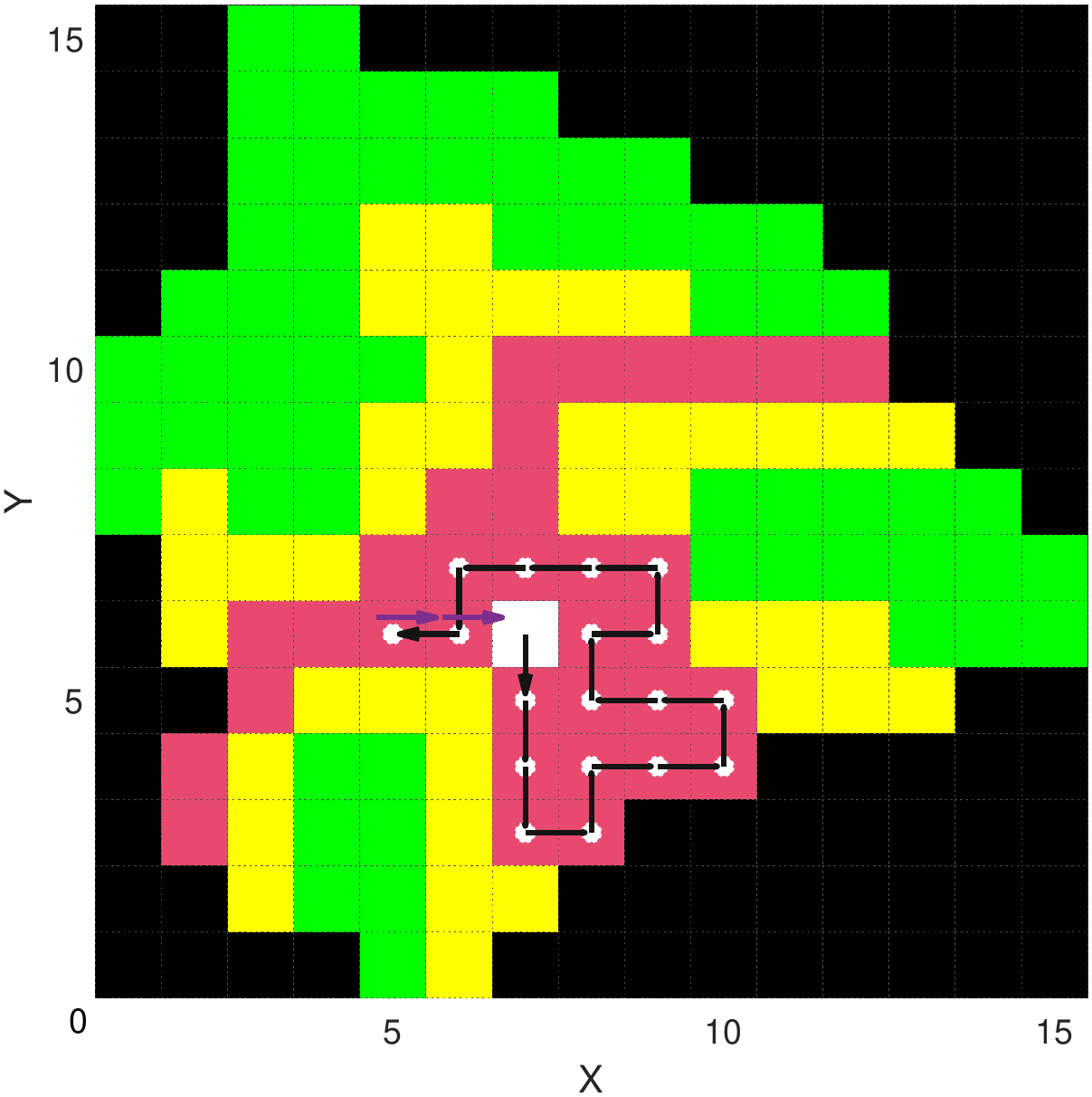}	
    \caption*{\textnormal{(b) $[x_\text{base}, y_\text{base}] = [7, 6]$, $B_\text{start} = 25, \ J_2 = 5$}}
	\end{subfigure}
	\vspace{5mm}
	\begin{subfigure}[b]{0.48\textwidth}
    \includegraphics[width=1\linewidth]{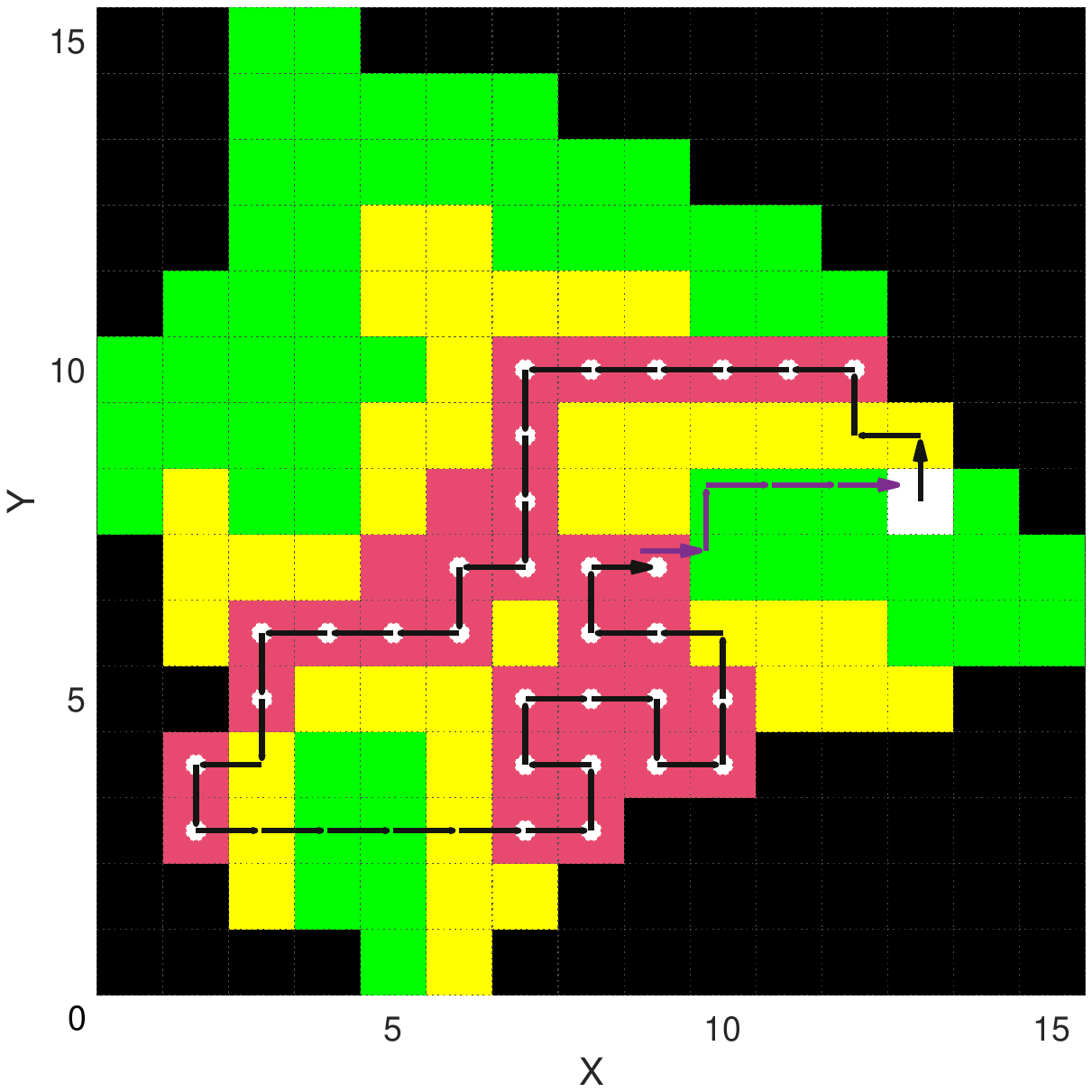}	
    \caption*{\textnormal{(c) $[x_\text{base}, y_\text{base}] = [13, 8]$, $B_\text{start} = 50, \ J_2 = 6$}}
	\end{subfigure}
	\hspace{5mm}
	\begin{subfigure}[b]{0.48\textwidth}
    \includegraphics[width=1\linewidth]{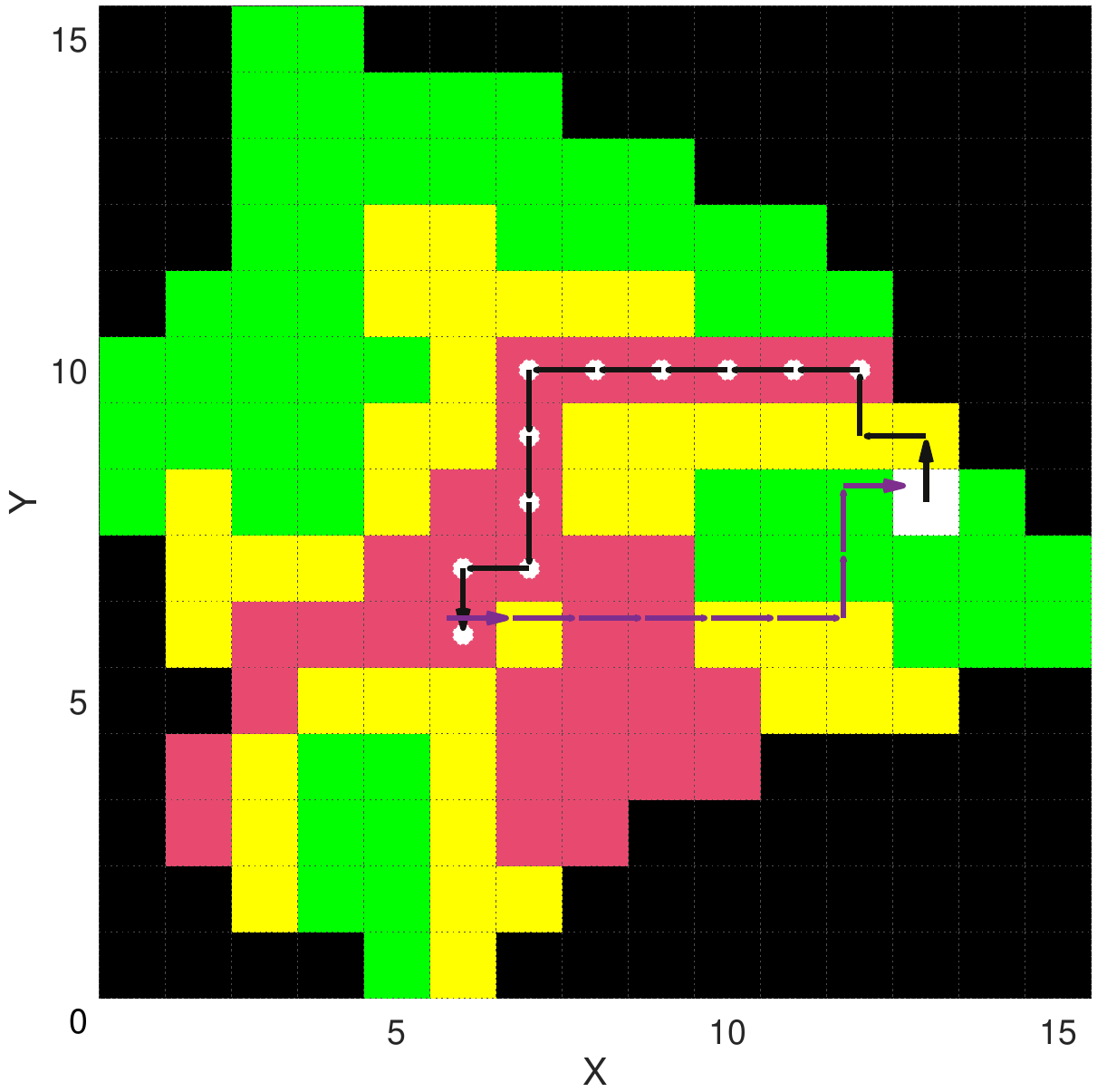}	
    \caption*{\textnormal{(d) $[x_\text{base}, y_\text{base}] = [13, 8]$, $B_\text{start} = 25, \ J_2 = 3$}}
	\end{subfigure}
    \caption{Application of the proposed algorithm on the Hawai`i island map in Fig.~\ref{Fig:Gridpath} for different initial locations and battery capacities. The red, yellow, green and blue cells denote the highest, medium, lowest and zero coverage importance, respectively. The white cell denotes the base location and the white dots show the next instantaneous location with the highest importance.  The black line denotes the outgoing trajectory, the purple line denotes the shortest-path return trajectory, and the arrowheads denote the directions.}
	\label{fig:HawaiiTraj}
\end{figure*}

We present the simulation results that show a series of trajectories chosen by Algorithm~\ref{alg1} in Fig.~\ref{fig:GaussianTraj} and ~\ref{fig:HawaiiTraj}. 
The trajectory consists of two segments, the optimal exploration segment from the base and then to the optimal return segment to the base.
The intermediate targets are chosen by comparing the number of cells that need to be visited to reach the position against the values assigned to the color of the cell and the penalty of revisiting a cell.
The simulation uses a $15 \times 15$ grid and creates two different types of importance maps with nonuniform reward as discussed below.
The trajectories were also generated on multiple maps by varying the base locations and battery capacities for performance analysis.

Fig.~\ref{fig:GaussianTraj} presents optimal trajectory planning on a map with cell coverage importance values following a multivariate Gaussian distribution.
For (a) and (b), the base is at $[x_\text{base}, \ y_\text{base}] = [7,6]$ with battery capacities 50 and 25 respectively and for (c) and (d), the base is at $[13,8]$ with the same battery capacities.
This map is relatively simple and the trajectories in all the sub-figures of Fig.~\ref{fig:GaussianTraj} use the available battery capacity to get to the higher reward locations and explore them efficiently before covering the areas of lower importance, which is shown by the black lines.
The optimal return path (purple lines) is calculated at each of the intermediate white dots (omitted to avoid cluttering) and the algorithm keeps exploring until it reaches the critical battery level, and finally returns to the base through the shortest path of highest reward from that particular cell as shown in the figures.

Similarly, Fig.~\ref{fig:HawaiiTraj} shows the performance of the proposed algorithm on a relatively complicated and real-life inspired map of (exaggerated) volcano flow on the island of Hawai`i. 
The coverage importance map is defined and overlaid on a square grid over the map of the island of Hawai`i as explained in Fig.~\ref{Fig:Gridpath}.
For (a) and (b), the base is at $[x_\text{base}, \ y_\text{base}] = [7,6]$, which corresponds to the peak of Mauna Lua, with battery capacities 50 and 25, respectively, and for (c) and (d), the base is at $[13,8]$, which corresponds to the general location of the city of Hilo, with the same battery capacities.
Even for this complicated map, the trajectories in all the sub-figures of Fig.~\ref{fig:HawaiiTraj} efficiently perform the exploration of the higher importance areas before covering the areas of lower importance (shown by the black trajectory).
The proposed algorithm performs generally well for both the maps as the UAV is always able to return to base and chooses a route that maximizes reward and minimizes revisiting penalties. 
The minimization of unspent energy varies depending on the geometry of cells of high reward and the path taken thus far. 

\begin{figure*}[]
    \centering
    \includegraphics[width=0.65\textwidth]{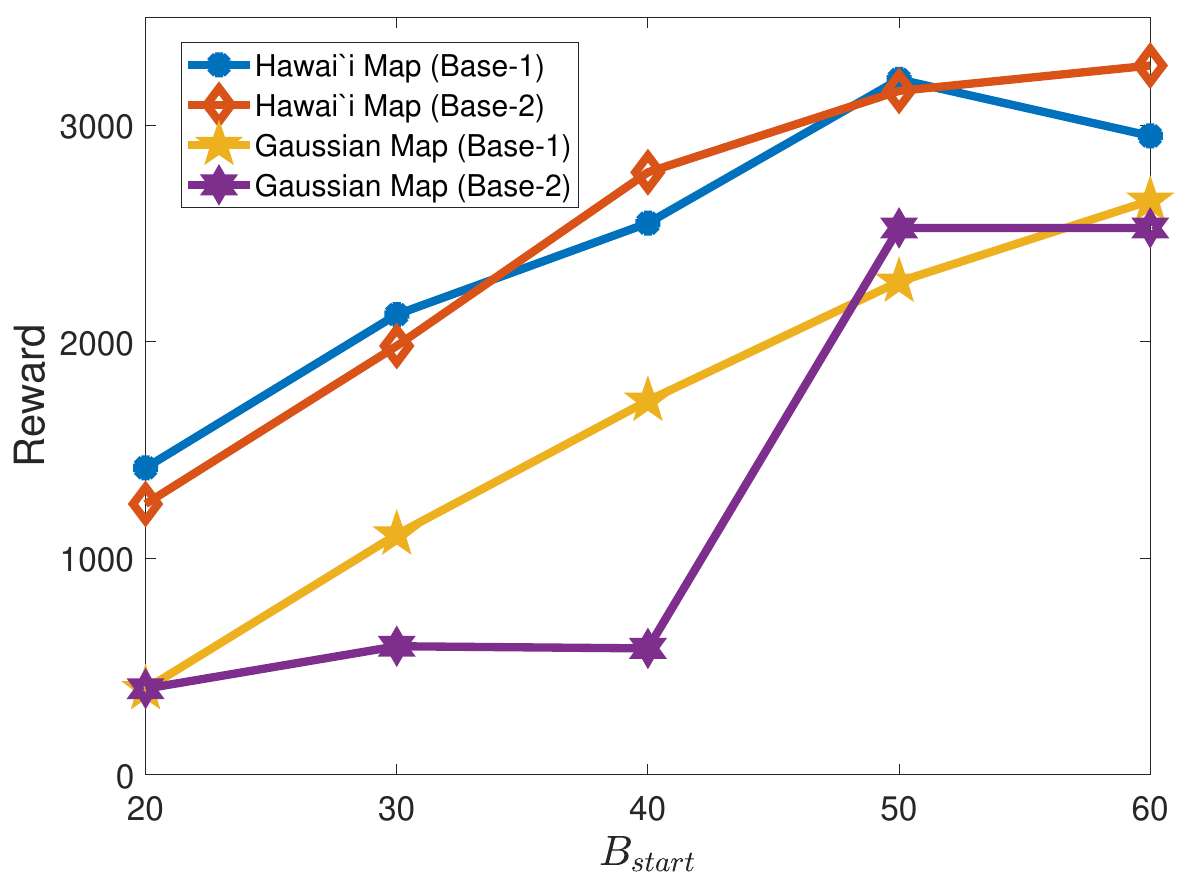} 
    \caption{Comparison of the trajectory performance in terms of the total reward collected for the maps and base locations seen in Figs~\ref{fig:GaussianTraj} and ~\ref{fig:HawaiiTraj} with by varying the initial battery capacities ($B_{\text{start}} = 20,30,40,50,60$). 
    The cell importance scores are set as $0, 1,10 \text{ and } 100$ (blue, green, yellow and red), and the penalty for revisiting a trial was $-100$ and an additional penalty of $-1$ for each unspent energy tile. Base 1 and Base 2 for both maps correspond to the initial locations $[7, 6] \text{ and } [13, 8]$, respectively.}
    \label{Fig:pwreff}
\end{figure*}

In Fig.~\ref{Fig:pwreff} we varied the starting battery capacities for all the maps and initial location scenarios in Figs.~\ref{fig:GaussianTraj} and \ref{fig:HawaiiTraj} to analyze the reward performance of the proposed algorithm. 
Since the Hawai`i island map has more higher importance cells, the total accumulated reward has higher values for both base locations for.
In addition, the overall performance is similar for the different initial locations and initial battery capacities, which shows that the trajectories generated are fairly consistent and optimal.
For the two probability maps the differences are most likely due to the clustering of the high importance cells in the Gaussian map. 
This is ultimately the result of only looking one step forward when determining a target and a respective return path. 
As we can see in Fig.~\ref{fig:GaussianTraj}(c), if there was additional energy, the next immediate step would cause the trajectory to circumnavigate the visited cells, which is also shown in Gaussian Map (Base-2) plot of Fig.~\ref{Fig:pwreff}, where there was not enough energy to reach the next target waypoint after the 30 energy run until 50 energy was provided. 
The algorithm then followed the same trajectory for the 40 energy trial as it did for the 30 energy and took the penalty.
The trajectory planning and reward performance results are in agreement with the algorithm expectations and show its applicability for a range of maps and scenarios.

\section{Conclusion}
\label{conclusion}

In this work we presented an algorithm that finds the optimal paths through a survey area with nonuniform importance to maximize the coverage and net coverage importance reward for a given initial battery capacity. We follow a set of dynamically generated trajectories dependent on the surrounding weights of the area. The UAV navigates through the area while avoiding the visited locations to maximize the exploration gain while being constrained by the minimum power to return to base. Further research will include dynamic targeting system that takes into account the weights of all possible points that can be visited in between targets as well as linking the trajectories that the UAV uses to the optimal energy required for the specific configuration and the total battery power allowance. We are also working on coordination of UAVs for exploration of a large area, where the sensing workload will be divided by an optimal mission coordination algorithm. Inclusion of wind (static and dynamic) over the deployment area and its effect on trajectory optimization is another important problem under investigation.

\section{Acknowledgments}

Snyder, Shriwastav, and Morrison-Fogel would like to thank Dr. Reza Ghorbani at the University of Hawai`i at M\=anoa for his insights and extensive discussions on this work. Song would like to thank the support from the National Science Foundation under awards CISE/IIS-2024928 and OIA-2032522.
\bibliography{Ref_coverage}

\begin{thebibliography}{14}
\newcommand{\enquote}[1]{``#1''}
\providecommand{\natexlab}[1]{#1}
\providecommand{\url}[1]{\texttt{#1}}
\providecommand{\urlprefix}{URL }
\expandafter\ifx\csname urlstyle\endcsname\relax
  \providecommand{\doi}[1]{\discretionary{}{}{}https://doi.org/#1}\else
  \providecommand{\doi}[1]{\discretionary{}{}{}\urlstyle{rm}\url{https://doi.org/#1}}\fi

\bibitem[{Sonmez et~al.(2015)Sonmez, Kocyigit, and Kugu}]{SonmezeKocyigitKugu}
Sonmez, A., Kocyigit, E., and Kugu, E., \enquote{Optimal path planning for UAVs
  using genetic algorithm,} \emph{Proceedings of the International Conference
  on Unmanned Aircraft Systems (ICUAS)}, Denver, CO, USA, 2015, pp. 50--55.

\bibitem[{Geiger et~al.(2006)Geiger, Horn, DeLullo, Niessner, and
  Long}]{GeigerHornDeLulloNiessnerLong}
Geiger, B., Horn, J., DeLullo, A., Niessner, A., and Long, L., \enquote{Optimal
  path planning of UAVs using direct collocation with nonlinear programming,}
  \emph{Proceedings of the AIAA Guidance, Navigation, and Control Conference
  and Exhibit}, Keystone, CO, USA, 2006, p. 6199.

\bibitem[{Sujit and Beard(2009)}]{SujitBeard}
Sujit, P.~B., and Beard, R., \enquote{Multiple UAV path planning using anytime
  algorithms,} \emph{Proceedings of the American Control Conference (ACC)},
  St.~Louis, MO, USA, 2009, pp. 2978--2983.

\bibitem[{Bortoff(2000)}]{Bortoff}
Bortoff, S.~A., \enquote{Path planning for UAVs,} \emph{Proceedings of the
  American Control Conference (ACC)}, Chicago, IL, USA, 2000, pp. 364--368.

\bibitem[{Dogan(2003)}]{Dogan}
Dogan, A., \enquote{Probabilistic path planning for UAVs,} \emph{Proceedings of
  the 2nd AIAA ``Unmanned Unlimited" Conference and Workshop \& Exhibit}, San
  Diego, CA, USA, 2003, p. 6552.

\bibitem[{Butenko et~al.(2013)Butenko, Murphey, and
  Pardalos}]{ButenkoMurpheyParadalos}
Butenko, S., Murphey, R., and Pardalos, P.~M., \emph{Cooperative Control:
  Models, Applications and Algorithms}, Vol.~1, Springer Science \& Business
  Media, 2013.

\bibitem[{Nikolos et~al.(2003)Nikolos, Valavanis, Tsourveloudis, and
  Kostaras}]{NikolosValavantisTsourveloudisKostaras}
Nikolos, I.~K., Valavanis, K.~P., Tsourveloudis, N.~C., and Kostaras, A.~N.,
  \enquote{Evolutionary algorithm based offline/online path planner for UAV
  navigation,} \emph{IEEE Transactions on Systems, Man, and Cybernetics, Part B
  (Cybernetics)}, Vol.~33, No.~6, 2003, pp. 898--912.

\bibitem[{Arantes et~al.(2016)Arantes, Arantes, Toledo, and
  Williams}]{ArantesArantesToledoWilliams}
Arantes, M. d.~S., Arantes, J. d.~S., Toledo, C. F.~M., and Williams, B.~C.,
  \enquote{A hybrid multi-population genetic algorithm for UAV path planning,}
  \emph{Proceedings of the Genetic and Evolutionary Computation Conference
  (GECCO)}, Denver, CO, USA, 2016, pp. 853--860.

\bibitem[{Li et~al.(2016)Li, Wang, and Sun}]{LiWangSun}
Li, D., Wang, X., and Sun, T., \enquote{Energy-optimal coverage path planning
  on topographic map for environment survey with unmanned aerial vehicles,}
  \emph{Electronics Letters}, Vol.~52, No.~9, 2016, pp. 699--701.

\bibitem[{Mittal and Deb(2007)}]{MittalDeb}
Mittal, S., and Deb, K., \enquote{Three-dimensional offline path planning for
  UAVs using multiobjective evolutionary algorithms,} \emph{Proceedings of the
  IEEE Congress on Evolutionary Computation}, Singapore, 2007, pp. 3195--3202.

\bibitem[{Dijkstra(1959)}]{dijkstra1959note}
Dijkstra, E.~W., \enquote{A note on two problems in connexion with graphs,}
  \emph{Numerische Mathematik}, Vol.~1, No.~1, 1959, pp. 269--271.

\bibitem[{Bellman(1958)}]{bellman1958routing}
Bellman, R., \enquote{On a routing problem,} \emph{Quarterly of Applied
  Mathematics}, Vol.~16, No.~1, 1958, pp. 87--90.

\bibitem[{Moore(1959)}]{moore1959shortest}
Moore, E.~F., \enquote{The shortest path through a maze,} \emph{Proceedings of
  the International Symposium on the Theory of Switching}, 1959, pp. 285--292.

\bibitem[{Ford~Jr(1956)}]{ford1956network}
Ford~Jr, L.~R., \enquote{Network flow theory,} Tech. rep., RAND Corporation,
  Santa Monica, CA, USA, 1956.

\end{thebibliography}

\end{document}